# Autonomous Dimension Reduction by Flattening Deformation of Data Manifold under an Intrinsic Deforming Field

Xiaodong Zhuang

Electronic Information College, Qingdao University, 266071 China

**Abstract**－A new dimension reduction (DR) method for data sets is proposed by autonomous deforming of data manifolds. The deformation is guided by the proposed deforming vector field, which is defined by two kinds of virtual interactions between data points. The flattening of data manifold is achieved as an emergent behavior under the elastic and repelling interactions between data points, meanwhile the topological structure of the manifold is preserved. To overcome the uneven sampling (or "short-cut edge") problem, the soft neighborhood is proposed, in which the neighbor degree is defined and adaptive interactions between neighbor points is implemented. The proposed method provides a novel geometric viewpoint on dimension reduction. Experimental results prove the effectiveness of the proposed method in dimension reduction, and implicit feature of data sets may also be revealed.

## 1. Introduction

To face the challenge of dramatically increasing data amount in applications, and also the "curse of data dimensionality", dimension reduction (DR) is an indispensable technique in practical data analysis [1-6]. Currently, non-linear DR methods is a research focus [5,6]. Manifold learning is one of the main non-linear methods. Besides non-linear feature extraction, manifold learning provides a geometric viewpoint for modern data analysis, where the data set is regarded as samples from a data manifold. Moreover, research has proved that the visual information represented in the neural system can also be modeled in a manifold-based way [7-9]. Therefore, manifold-based methods have attracted extensive research attention since the publication of the Isometric Mapping and Local Linear Embedding methods. Many other methods have been proposed such as Laplacian Eigenmaps, Maximum Variance Unfolding, Riemannian Manifold Learning, Locally Linear Coordination, Stochastic Neighbor Embedding, Local Tangent Space Alignment, etc. [10-17]. Moreover, some frameworks of manifold learning have also been proposed to classify various manifold learning methods, such as Graph Embedding Framework, Patch Alignment Framework, and Kernel Framework .

  The existing methods usually have the form of solving optimization or linear programming under certain constraints. Although these learning methods have achieved impressive experimental results, some common problems still exist such as the difficulty of intrinsic dimension estimation and neighbor selection [21-25]. Moreover, it has been pointed out that current manifold learning methods may fail because of the extremely high dimension or high local curvature of the data manifold [21-23]. Another problem may also invalidate manifold learning that the practical data sets usually do not satisfy the ideal precondition of sufficiently dense and uniform sampling on the manifold. These become the bottlenecks for practical application of manifold learning.

  In order to overcome such difficulties, in this paper a novel DR method is proposed from the geometric viewpoint, which interprets the DR process as the "flattening" of the manifold in the embedding space. Natural intelligent mechanism is usually self-organized with an emergent behavior of swarm with simple individuals (such as ant colony, bee colony, neuron systems, etc.) [26-28]. The proposed method implements such emergent behavior mechanism, in which the data manifold (in a discrete form) deforms in an autonomous self-evolution way under the virtual interaction between the data points. The DR result can be naturally derived from the flattened manifold. Moreover, in the proposed method, "soft neighborhood" of data point is presented to overcome the difficulties of neighbor selection and non-uniform sampling. And the deforming result can naturally indicate the intrinsic dimension of the manifold. The experimental results prove the effectiveness of the proposed method, which provides a new direction to design manifold learning methods.

## 2. Method Overview

The proposed method introduces two virtual interactions between data points, which cause the deformation of the data manifold. By proper design of the interactions, the flattening of the manifold occurs as an emergence effect. The design of the proposed method is inspired by the viewpoint of geometrically interpreting dimension reduction as the flattening of data manifold.

From the geometric viewpoint, DR technique produce the "flattening" effect on the manifold geometry. New idea is inspired by analyzing the inverse process of flattening, i.e. how a flat manifold be changed to a geometry of more complex shape in the embedded space $R^n$. For intuitive comprehension, consider the following cases. Folding or curling a piece of paper will make the points on the paper leave the initial plane, and move to another position in $R^3$. In another case, making a flat elastic film uneven or cratering can also cause some points move to their new position in $R^3$. For the first case, in the folding or curling process, points that are originally far away may come close. To restore the original shape, the distance between such points should be increased as large as possible, while preserving the distance between those neighbor points that are close enough (otherwise the paper will be torn). For the second case, because the geometric structure in the convex or concave parts are already nonlinear, optimal DR results may be obtained by proper "stretching". Overall consideration of these cases inspires a reasonable way of DR by "flattening", in which the distances between points are increased as possible, while properly preserving the distance between those points that are originally close enough. Similar idea appeared in the Maximum Variance Unfolding (MVU) method, but in MVU the DR is achieved in traditional optimization framework.

In this paper, a novel type of DR method is proposed that the manifold geometry is autonomously "flattened" in the embedded space by virtual interactions between the points on the manifold. Especially, the soft neighborhood is proposed to overcome the problem of non-even sampling on manifold, which may invalidate traditional methods including MVU. By the emergent behavior of the interacting manifold points, the manifold can autonomously deform (self-evolution), and the DR results can naturally be achieved by the deforming result of the manifold.

## 3. The Soft Neighborhood

Useful information and intrinsic features are implicit in contained in the topological structure of the data manifold. Preserving the topological structure is a basic constraint for manifold learning methods. The discrete data points are considered as samplings from the data manifold, and the topological structure of the data manifold is expressed by the neighborhood relationship between data points. In manifold learning, it is usually the first step to search the neighbors for each data point. Current methods include fixed neighborhood-radius or fixed number of neighbor points, which are suitable for uniform and dense sampling of the manifold. However, practical data sets may have limited number of data points, and the sampling is often non-uniform. This will cause some misjudgement of neighbor point, which may invalidate the learning. For the *k* nearest neighbor points, if *k* is large, some non-neighbor point will be included, which will cause "short-circuit edge". To overcome this problem, the "soft neighborhood" method is proposed.

For each point $p_i$ on the initial manifold (i.e. the one before deforming), find the *k* nearest points as its neighbor set $N_i=\{q_1, q_2, \ldots, q_m\}$. The value of *k* can be properly large in order that no true neighbor point is missed. To overcome the "short-circuit edge" problem, the neighbor degree is defined. Let $d_{ij}$ denote the distance between $p_i$ and $q_j$. Let $d_{min}$ denote the minimum value in $\{d_{ij}\}$, $j=1,2,\ldots,m$. The neighbor degree for $p_i$'s neighbor set is:

$$ND_{ij}=d_{min}/d_{ij} \qquad j=1,2,\ldots,m \tag{1}$$

where $ND_{ij}$ is the neighbor degree of $q_j$ to $p_i$. Obviously, there is $0<ND_{ij}<=1.0$, which is similar to the degree of membership in a fuzzy set [29]. If $q_j$ is actually a non-neighbor point (a misjudged one), $ND_{ij}$ value will be very small. Therefore, $ND_{ij}$ quantitatively expresses to what degree $q_j$ is a true neighbor point of $p_i$. Proper use of $ND_{ij}$ in the DR technique can eliminate the severe interference of "short-circuit edge". In the proposed learning method, the value of $ND_{ij}$ plays an important role in the definition of interactions between data points. A more precise expression of $p_i$'s soft neighborhood is:

$$SN_i=\{(q_j, ND_{ij})\} \qquad j=1,2,\ldots,m \tag{2}$$

where $q_j$ belongs to the *k*-nearest points, and $ND_{ij}$ is the neighbor degree of $q_j$ to $p_i$.

## 4. Intrinsic Deforming Vector Field with Flattening Effect

In order to implement dimension reduction by manifold deforming, two virtual interactions between data points are proposed to derive an autonomous deforming process. They are the repelling and elastic interactions. The repelling vector from the data point $p_j$ to $p_i$ is defined as:

$$\vec{V}_{ij}^r = \begin{cases} \frac{(1.0-ND_{ij})\cdot(\vec{p}_i-\vec{p}_j)}{d_{ij}} & p_j \in N_i \\ \frac{\vec{p}_i-\vec{p}_j}{d_{ij}} & otherwise \end{cases} \tag{3}$$

where $\vec{p}_i$ and $\vec{p}_j$ are the position vectors of data points $p_i$ and $p_j$ in $R^n$. $d_{ij}$ is the distance between the two points in the deforming process. $ND_{ij}$ is the neighboring degree of $p_j$ to $p_i$. Because the vector $(\vec{p}_i - \vec{p}_j)$ points from $p_j$ to $p_i$, if $p_i$ moves along this direction it will move away from $p_j$. Therefore the vector defined in Equation (3) has a repelling effect between $p_i$ and $p_j$.

On the other hand, the elastic interaction vector between $p_i$ and $p_j$ is defined as:

$$\vec{V}_{ij}^e = \begin{cases} \frac{ND_{ij}\cdot(d_{ij}^0-d_{ij})\cdot(\vec{p}_i-\vec{p}_j)}{d_{ij}} & p_j \in N_i \\ 0 & otherwise \end{cases} \tag{4}$$

where $d_{ij}^0$ is the Euclidean distance between $p_i$ and $p_j$ on the original manifold (before deforming). $d_{ij}$ is the distance between $p_i$ and $p_j$ in the deforming process. Therefore, the interaction vector defined in Equation (4) will alter according to the manifold shape in the deforming process. For point $p_i$, this elastic interaction only exists for the points in the soft neighborhood $NS_i$. For point $p_j$ in $NS_i$, if it comes closer to $p_i$ in deforming (i.e. the current distance $d_{ij}$ is smaller than the original value $d_{ij}^0$), it will repel $p_i$, otherwise it will attract $p_i$. This is just an elastic effect which functions as preserving the distance between neighbor points (i.e. keep the neighbor structure in the deforming process).

In Equation (3) and (4), $ND_{ij}$ properly weights the two different interactions between the data points in a soft neighborhood. In case there is "short-circuit edge", the actual non-neighbor point $p_j$ will have very small $ND_{ij}$ value (i.e. close to zero). And the interaction between $p_j$ and $p_i$ is mainly repelling, just similar to those non-neighbor points. Therefore, the problem caused by "short-circuit edge" can be solved in an adaptive way.

The total interaction on $p_i$ from all the other data points is the weighted sum of the above two interactions:

$$\vec{V}_i = \alpha_1 \cdot \sum_{\substack{j=1 \\ j \neq i}}^{N} \vec{V}_{ij}^{r} + \alpha_2 \cdot \sum_{\substack{j=1 \\ j \neq i}}^{N} \vec{V}_{ij}^{e} \qquad (5)$$

where $N$ is the number of data points. $\alpha_1$ and $\alpha_2$ are two weight coefficients that balance the two kinds of interactions. $\vec{V}_i$ is defined as the deforming vector on $p_i$. Because $\vec{V}_i$ is entirely determined by the current position of all data points (i.e. the manifold itself), and no external influence is involved, this vector field on the manifold is intrinsic. If each $p_i$ moves according to $\vec{V}_i$ (i.e. take $\vec{V}_i$ as the displacement vector), one step of manifold deforming will happen. Moreover, if the step repeats, the deformation of data manifold will proceed step by step. Due to the intrinsic nature of the proposed vector field, the deformation under it is a kind of self-evolution of the manifold. With the two different kinds of interactions in Equation (3) and (4), the deforming process will converge to a "flattened" result, which can naturally derive the dimension reduction result.

## 5. The Algorithm

Based on the above definitions, the proposed DR algorithm is as follows.

**Step1**: Calculate the Euclidean distance $d_{ij}$ between each pair of data points in the data set.

**Step2**: Find the $k$ nearest neighbor points for each data point as its soft neighborhood point set.

**Step3**: For each data point $p_i$, calculate the neighbor degree $ND_{ij}$ of each point in its soft neighborhood set.

**Step4**: Initialize the count of deforming steps $C$ as zero.

**Step5**: For each data point $p_i$, calculate the current displacement vector $\vec{V}_i$ according to the current position of each point (i.e. the current manifold shape) in $R^n$.

**Step6**: Update the position of each point $p_i$ with the displacement vector $\vec{V}_i$

**Step7**: Increase $C$ by 1.

**Step8**: Check whether the termination condition is satisfied (the sum of each point's displacement is smaller than a threshold $\varepsilon$, or $C$ reaches a given value). If not, return to Step 5. Otherwise, go to **Step 9**.

**Step9**: Perform Principle Component Analysis (PCA) on the flattened manifold, and obtain the final dimension reduction result (the estimated intrinsic dimension of manifold is the number of principle components, and the low-dimension coordinate of each data point is the projection on the principle component vectors).

The above algorithm first flattens the manifold in $R^n$, and then PCA is used to extract manifold dimension and the DR result, because the manifold has already deformed to a fairly flat geometry.

It must be noted that the proposed method belongs to global learning considering the definition of $\vec{V}_{ij}^{r}$, and the progressive (or stepwise) spread of local deformation to distant areas on the manifold. Although the elastic interaction is defined within a soft neighborhood of a point, based on the connectivity of any two points in the manifold topology, this local interaction will gradually affects the points far away. The local-global interaction of data points results in the autonomous deforming of the manifold, or its self-evolution. Although PCA is used in the last step of the algorithm, the overall method is non-linear due to the manifold deformation.

# 6. Experiment Results and Analysis

The proposed method is implemented by programming simulation. In the preliminary experiments, it is discovered that the values of $\alpha_1$ and $\alpha_2$ have obvious impact on the results. If $\alpha_1$ is much larger, the repelling between data points will be very strong. The manifold will be rapidly flattened and also stretched, but the distance between neighbor points can hardly be preserved. (Interestingly, in this case the deforming will still reach a balance state, in which the elastic interaction between neighbor points becomes strong attraction to counteract the repelling between points.) On the contrary, if $\alpha_2$ is much larger, the elastic interaction between neighbor points will be strong, which can preserve the neighbor distances well. But the deforming of the manifold will become very slow, or even terminated before the manifold is sufficiently flattened.

A dynamic alternation strategy is proposed to overcome the above problem. In the deforming process, let $\alpha_1$ rises and falls periodically, but keep $\alpha_2$ as constant. In this way, the repelling will prevail over the elastic interaction for some time, and then the elastic interaction will in turn become dominant. Correspondingly, the deforming process will alternates periodically between the two stage of "flattening" and "restoring neighbor distance", which implements dynamic balancing between the two interactions. In the experiments, $\alpha_1$ and $\alpha_2$ are defined as:

$$\alpha_1 = 10^{-4} \cdot \cos\left(\frac{2\cdot\pi\cdot mod(C,T)}{T}\right) \qquad T = 60$$

$$\alpha_2 = 0.1 \tag{6}$$

where $C$ is the count of deforming steps. $T$ is the period of $\alpha_1$'s oscillation. *mod* is the complementation (or remainder) operation.

Therefore, **Step 5** in the algorithm should include the updating of $\alpha_1$ and $\alpha_2$ according to Equation (6) before the calculation of the displacement vector. The experimental results prove the effectiveness of this dynamic alternation strategy.

Experiments have been done on simple test data sets, and also practical data sets in real world applications. The results and analysis are as follows.

## 6.1 Experiments on test data sets

Preliminary experiments have been done for typical types of surfaces in $R^3$. Some results are shown in Fig. 1 to Fig. 9 for the half cylinder side face and the Gaussian surface (corresponding to the curling and convex or concave cases respectively).

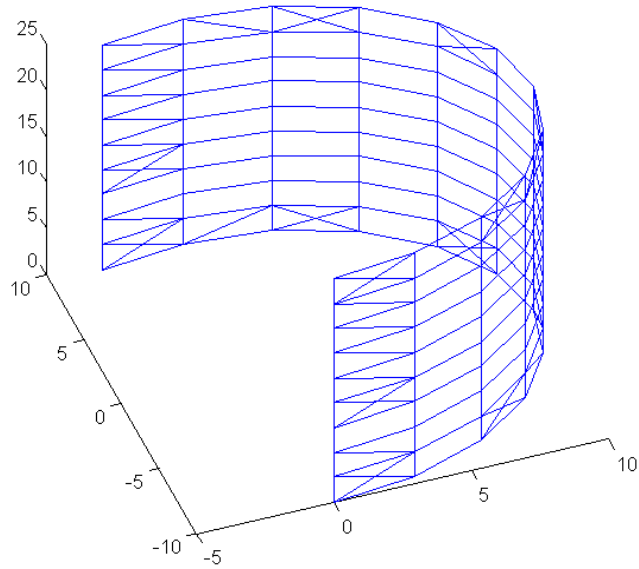

Fig. 1 The mesh of the discrete cylinder side face

Fig. 1 shows the mesh of the discrete cylinder side face, which has 120 data points. Because the neighborhood relationship is the basis of manifold topology structure, the DR results consist of nodes for data points, and edges for the representation of neighborhood relationship. The nodes represent the data points, and each edge connects two neighbor points in the data set.

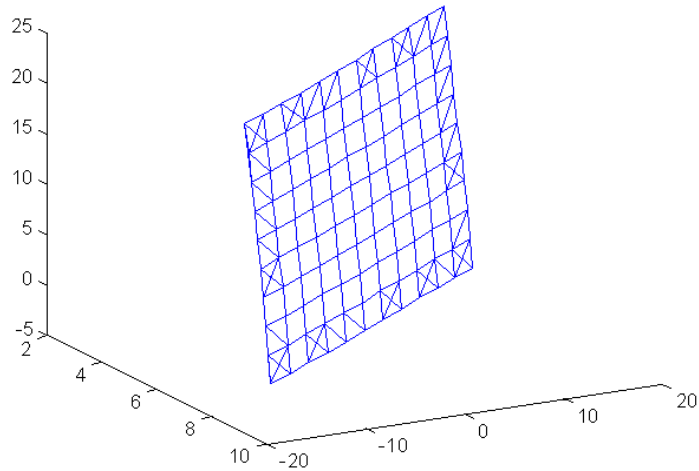

Fig. 2 The deforming result of Fig. 1 in $R^3$

Fig. 2 shows the deforming result of Fig. 1 in $R^3$. In the deforming result, the cylinder side face is totally flattened. Fig. 3 shows the final dimension reduction result in $R^2$, which is the result of PCA on Fig. 2. In Fig. 3, each node point represents the data after dimension reduction, and each edge connects two data points which correspond to neighbor points in the original data set. Moreover, the length of the edge represents the Euclidean distance between the two neighbor data points. The nodes in Fig. 3 are numbered, which may facilitate further analysis. In this way, the dimension reduction result and the topology structure of data can be clearly demonstrated. In Fig. 3, the data points are evenly distributed in $R^2$, which corresponds to the evenly sampling of the cylinder side face shown in Fig. 1. It proves the effectiveness of the proposed DR method on the type of curling surface.

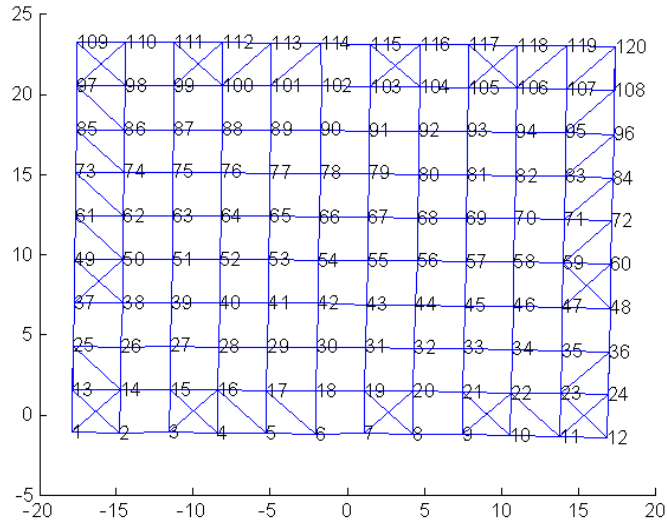

Fig. 3 The final dimension reduction result of Fig. 1 in $R^2$

Fig. 4 to Fig. 9 show the experimental results for the Gaussian surfaces. Fig. 4 and Fig. 7 show two Gaussian surfaces with different variance values, which have 120 data points respectively. Fig. 7 has a smaller variance value, therefore the shape appears much sharper. Fig. 5 and Fig. 8 are the deforming results in $R^3$ respectively. The final DR results after PCA are shown in Fig. 6 and Fig. 9. The intrinsic dimension of the data sets can be clearly revealed as two. Due to the non-linear property of the Gaussian surface, the dimension reduction results are not evenly distributed. However, the distance between neighbor points are relatively preserved as possible.

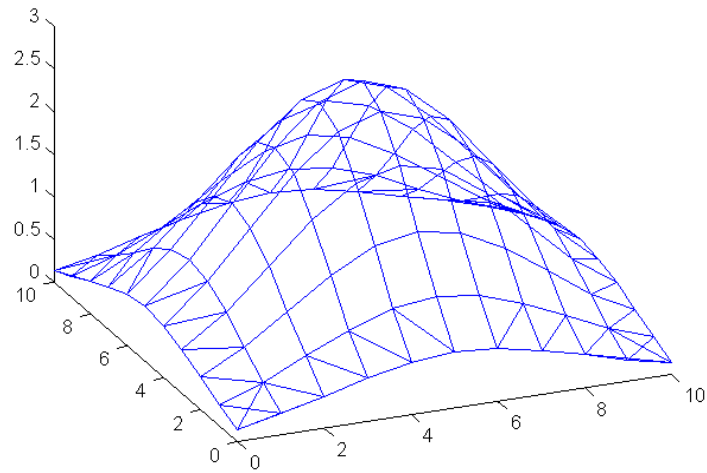

Fig. 4 The Gaussian surface with variance value 6

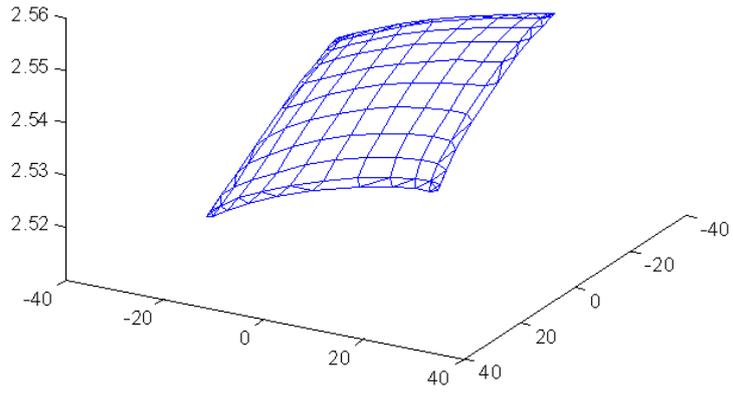

Fig. 5 The deforming result of Fig. 4 in $R^3$

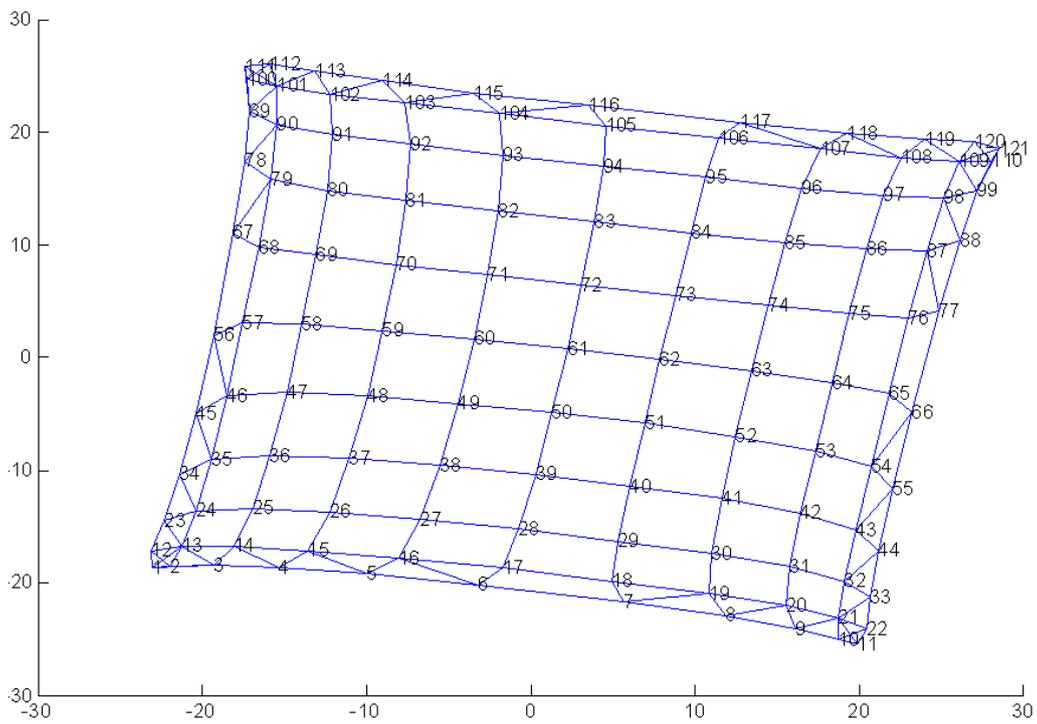

Fig. 6 The final dimension reduction result of Fig. 4 in $R^2$

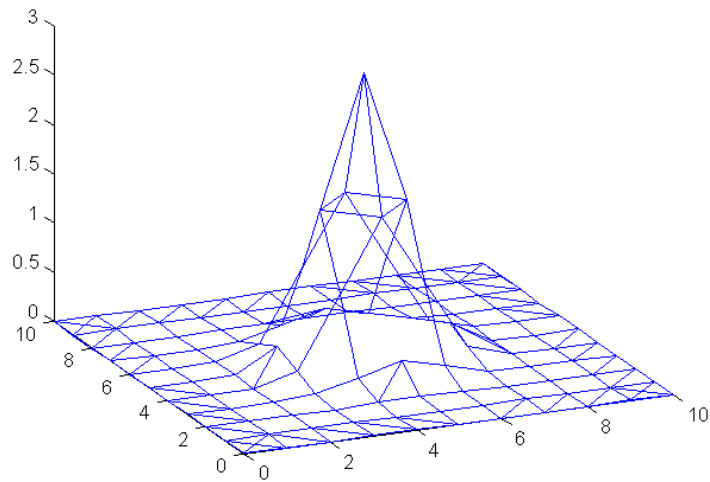

Fig. 7 The Gaussian surface with variance value 2

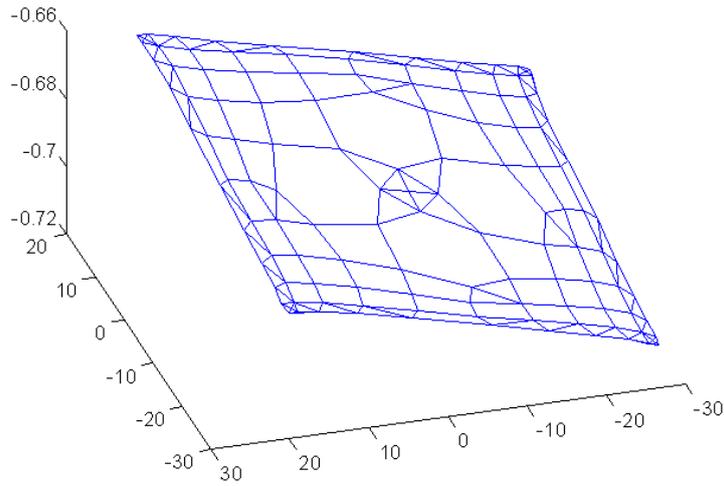

Fig. 8 The deforming result of Fig. 7 in $R^3$

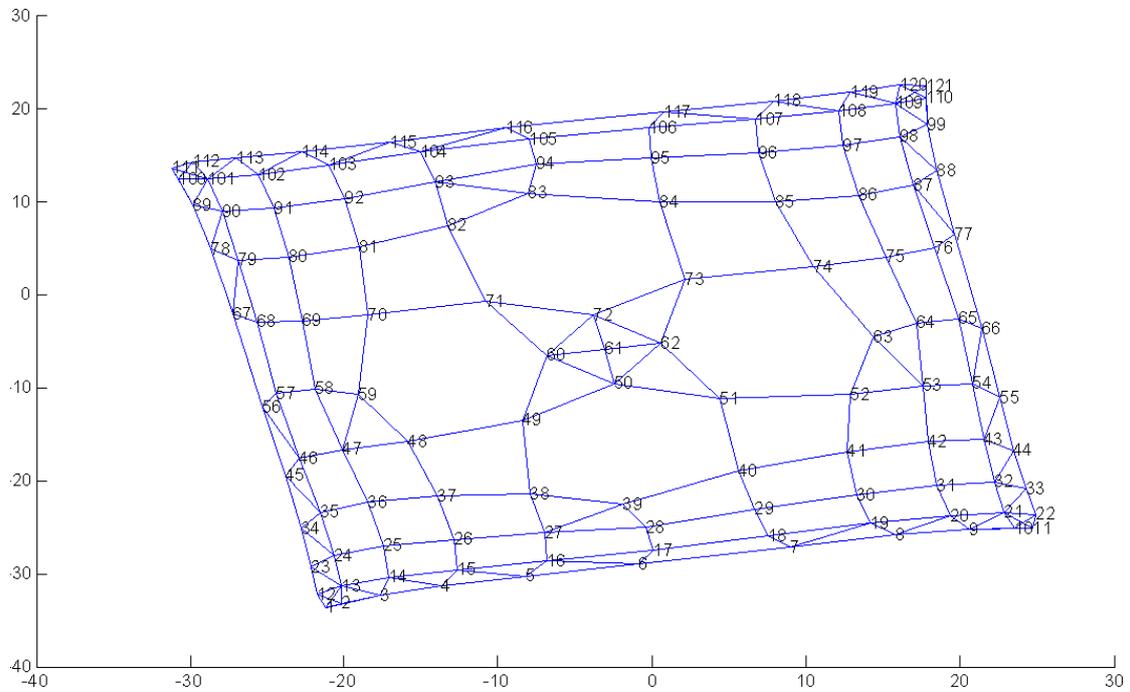

Fig. 9 The final dimension reduction result of Fig. 7 in $R^2$

## 6.2 Experiments on practical data sets in applications

Fig. 10 and Fig. 11 show the experimental results on an image set from the "Object Pose Estimation Database" [30]. This image set of auto fuse is captured under different horizontal and vertical viewpoints. The image data set is from Internet [31]. Fig. 10 shows the data set with a number assigned to each image. The dimension reduction result is shown in Fig. 11. The intrinsic dimension of this image set is estimated as two. Each node point in Fig. 11 represents an image with the same number in Fig. 10. The edges connect the pairs of nodes corresponding to the neighbor data points in the original data set. The result reveals that the images change along two different dimensions. The first dimension is the *x*-axis in Fig. 11, which corresponds to the change of horizontal viewpoint. The second one is the *y*-axis in Fig. 11, which corresponds to the change of vertical viewpoint. It should be noted that the points at the lower right corner in Fig. 11 are much closer, because the method preserves the distance between neighbor points, and those distances are relatively small in the original image data set. Therefore, the topology structure of the image set is effectively extracted and represented by the result in Fig. 11.

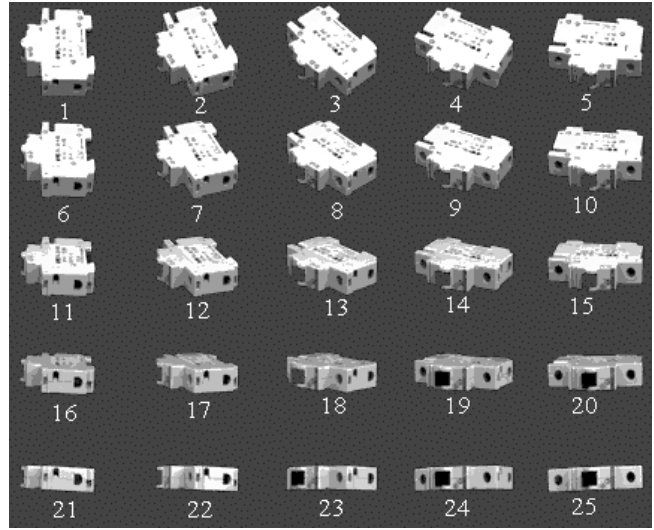

Fig. 10 The image set of auto fuse

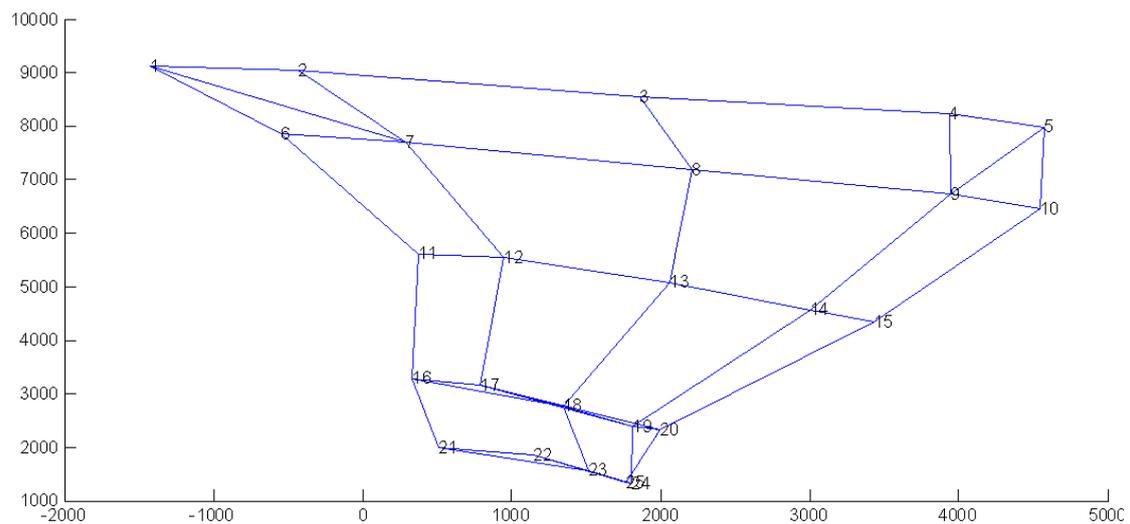

Fig. 11 The final dimension reduction result of the image set of auto fuse in $R^2$

Fig. 12 shows a group of image sequence from the "Columbia Object Image Library (COIL-20)", which is captured for rotating and simultaneously resizing objects [32]. In Fig. 12, the toy rotates for 360 degrees, together with simultaneous size variation. The image data set is from Internet [33]. The final result of dimension reduction is shown in Fig. 13, where each node is labeled with the corresponding image number. The toy images are also displayed near their corresponding nodes. Fig. 13 shows a closed curve representing the rotating angle variation from 0 to 360 degrees. The $x$-axis can be interpreted as the left-right rotating angle. The $y$-axis is related to the size factor. In Fig. 13, the points are very close at the top, lower left and lower right areas on the curve (indicated as area A,B and C in the figure). Therefore, these three areas are shown in Fig. 14, 15, 16 in more details. The variation of the images along the curve in Fig. 13 is consistent with the rotating process, and also consistent with the resizing process. Therefore, Fig. 13 can represent the topology structure of this data set.

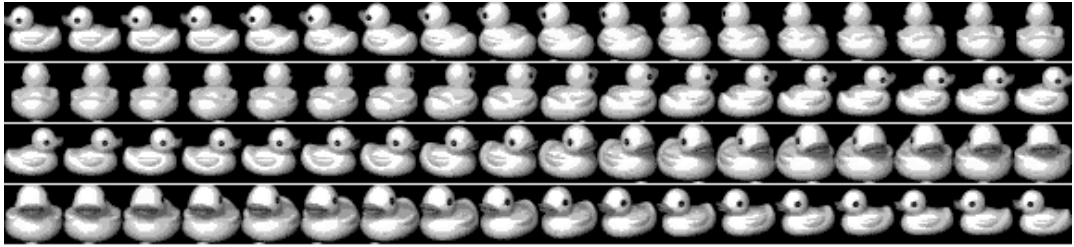

Fig. 12 The toy image sequence in the "Columbia Object Image Library (COIL-20)"

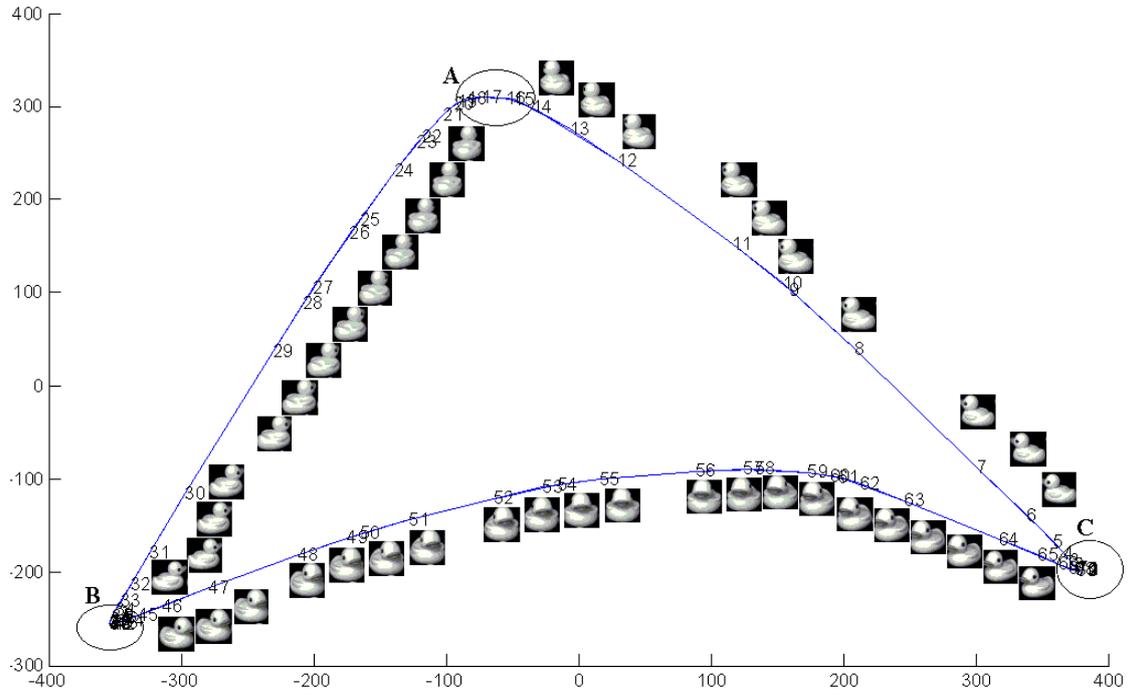

Fig. 13 The final dimension reduction result of the toy image sequence in $R^2$

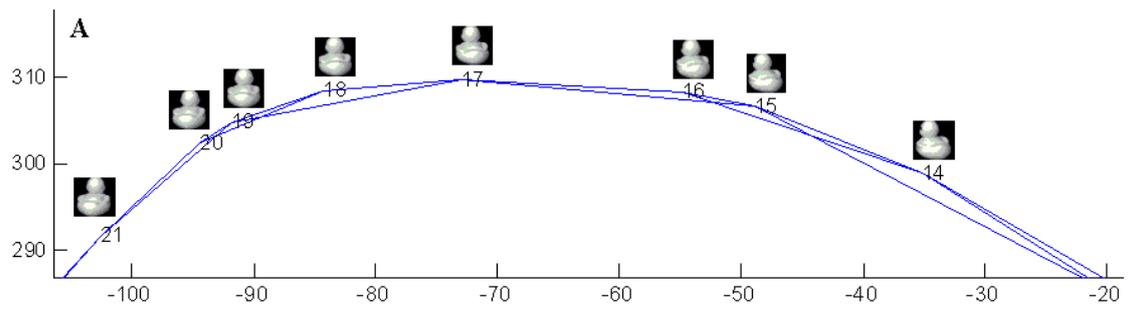

Fig. 14 The detailed demonstration of area A in Fig. 13

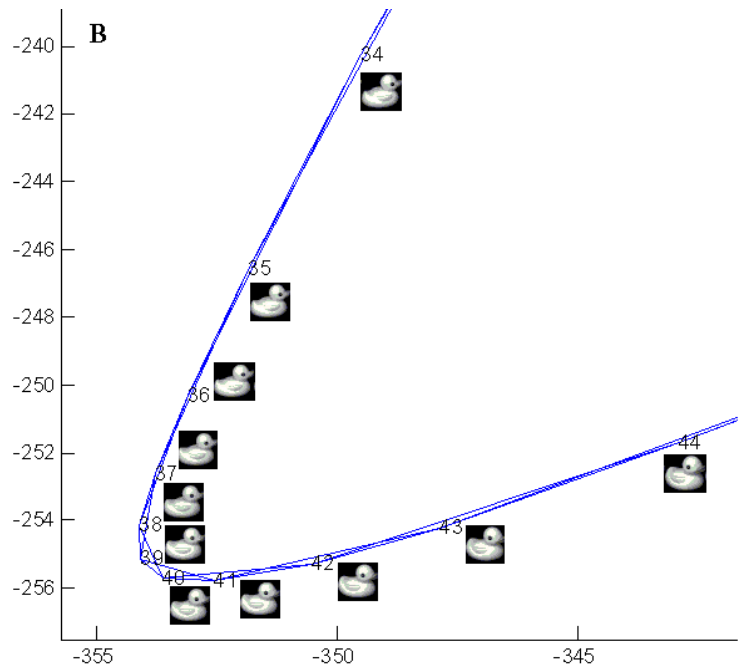

Fig. 15 The detailed demonstration of area B in Fig. 13

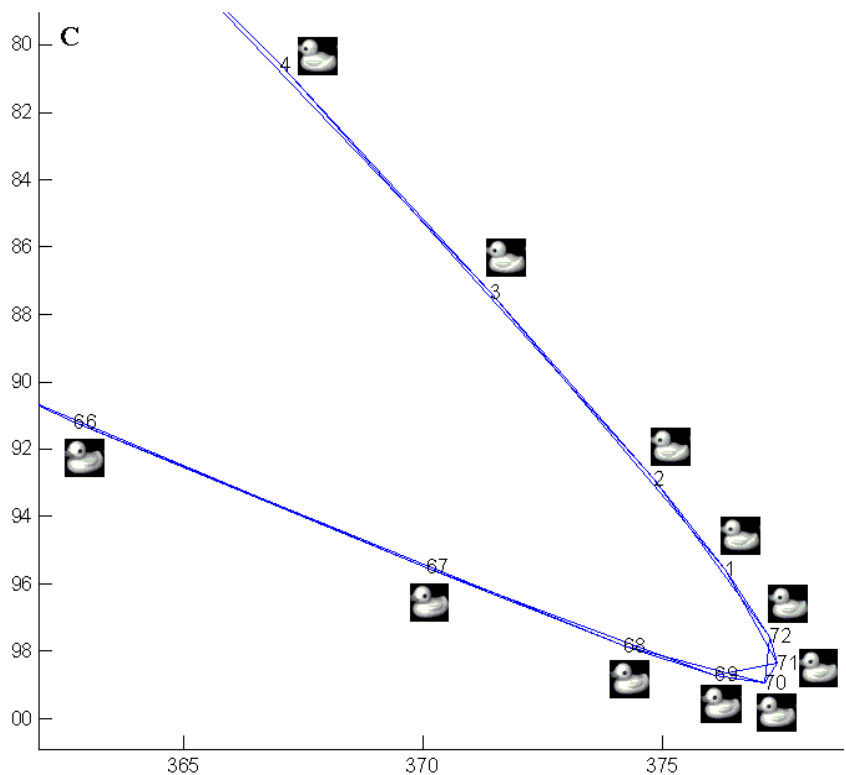

Fig. 16 The detailed demonstration of area C in Fig. 13

Moreover, in order to investigate the intrinsic dimension estimation, in the experiment the intermediate result after each deforming step is analyzed by PCA, and the relative ratios of the primary components are recorded. The variation of such ratios for the six most significant components are shown in Fig. 17. The six curves in Fig. 17 are labeled with the numbers of component order. Fig. 17 indicates that the most significant and second significant components have increasing proportion rates. But the ratios of the other four components decrease obviously with the deformation going on. This reveals the data set has two main latent variables, which just accords with the rotating angle and size factors. The fluctuation of the curves in Fig. 17 is due to the dynamic periodic adjustment of $\alpha_1$ and $\alpha_2$ in Equation (6).

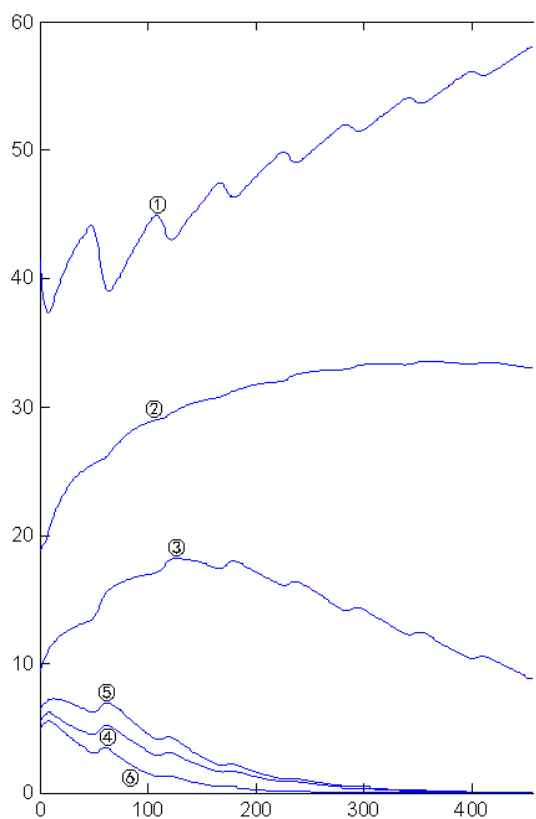

Fig. 17 The relative ratios of the first 6 primary components in the deforming process

Experiments have been done on the "Extended Yale Face Database B", which are groups of face images captured for different people [34]. The person and the camera keep motionless, but the illumination intensity and angle change. The image data set is from Internet [35]. Fig. 18 shows one group of images for a person. Fig. 19 shows the final result of dimension reduction. The node points in Fig.19 represents the images, which is labeled with their corresponding numbers. The face images are also displayed near the corresponding nodes. The edges connects the neighbor points. In the result, the *x*-axis in Fig. 19 can be interpreted as the illumination intensity factor, and the *y*-axis represents the illumination angle. For quantitative analysis, the sum of pixel intensity is calculated for each image as the representation of illumination intensity. And the intensity difference between left and right half of each image is also calculated as the representation of illumination angle factor. Fig. 20 shows the distribution of intensity summation along the *x*-axis, which indicates that the illumination intensity has increasing tendency along the *x*-axis. Fig. 21 shows the left-right difference of intensity, which indicates that the illumination angle changes from right to left along the *y*-axis. Therefore, the DR result clearly reveals the two factors underlying the photo images.

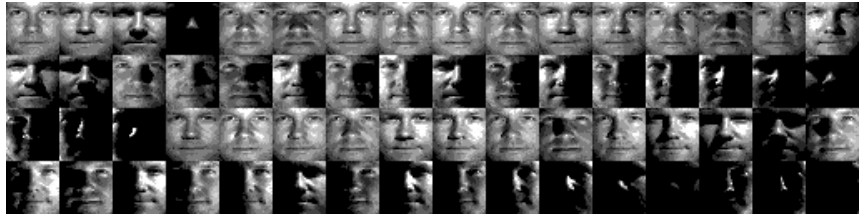

Fig. 18 One group of images for a person in the "Extended Yale Face Database B"

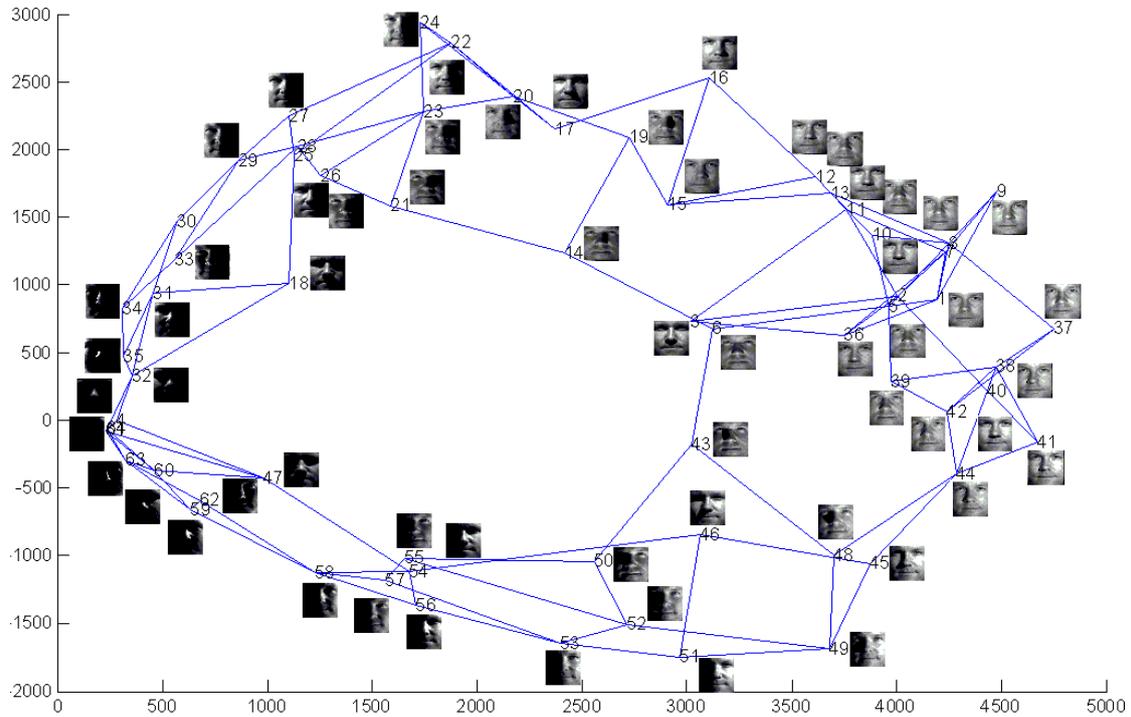

Fig. 19 The final dimension reduction result for the image sequence in Fig. 18

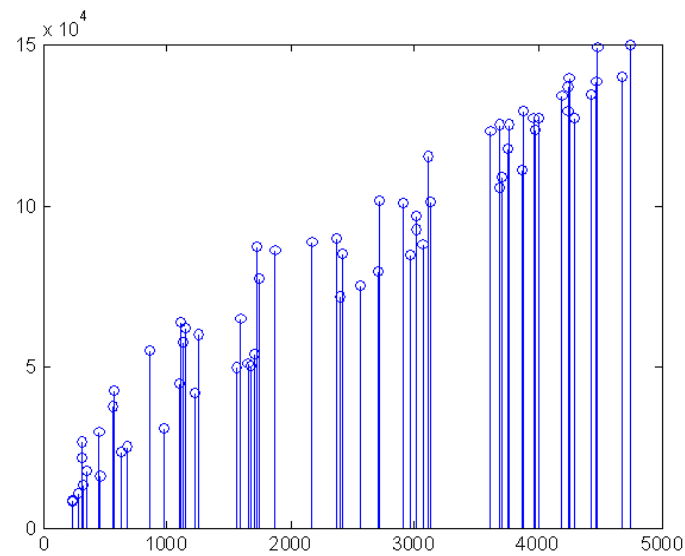

Fig. 20 The intensity summation for each image along the *x*-axis in Fig. 19

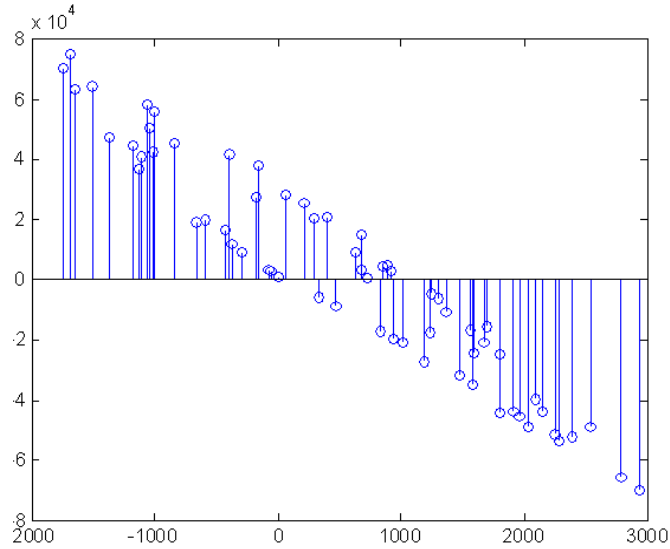

Fig. 21 The left-right difference of intensity for each image along the *y*-axis in Fig. 19

Experiments have been done for the cropped images of the "UMIST Face Database", which includes image sequences captured when people turn their heads[36]. The image data set is from Internet [37]. Fig. 22 shows one group of the images. Fig. 23 shows the final DR result, where the nodes are labeled with numbers. The corresponding images are also displayed. The *x*-axis represents the major dimension, which reflects the angle between the head's orientation and the posteroanterior direction. Interestingly, there exists non-negligible second dimension in the result, which reflects the angle between the head's orientation and the 45 degree direction of the face. In Fig. 23 this dimension reaches its minimum at node point 17 (i.e. the 17th image in the sequence), which is the one closest to the visual angle of 45 degree. On the other hand, long-term practice and experience in photography indicate that the most suitable angle of view to demonstrate the 3D effect of faces or objects is 45 degree. This indicates that the proposed method has the potential to reveal implicit features of data sets.

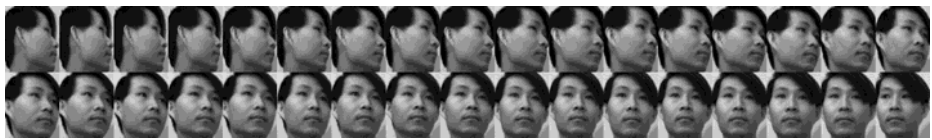

Fig. 22 One group of the images from the cropped images of the "UMIST Face Database"

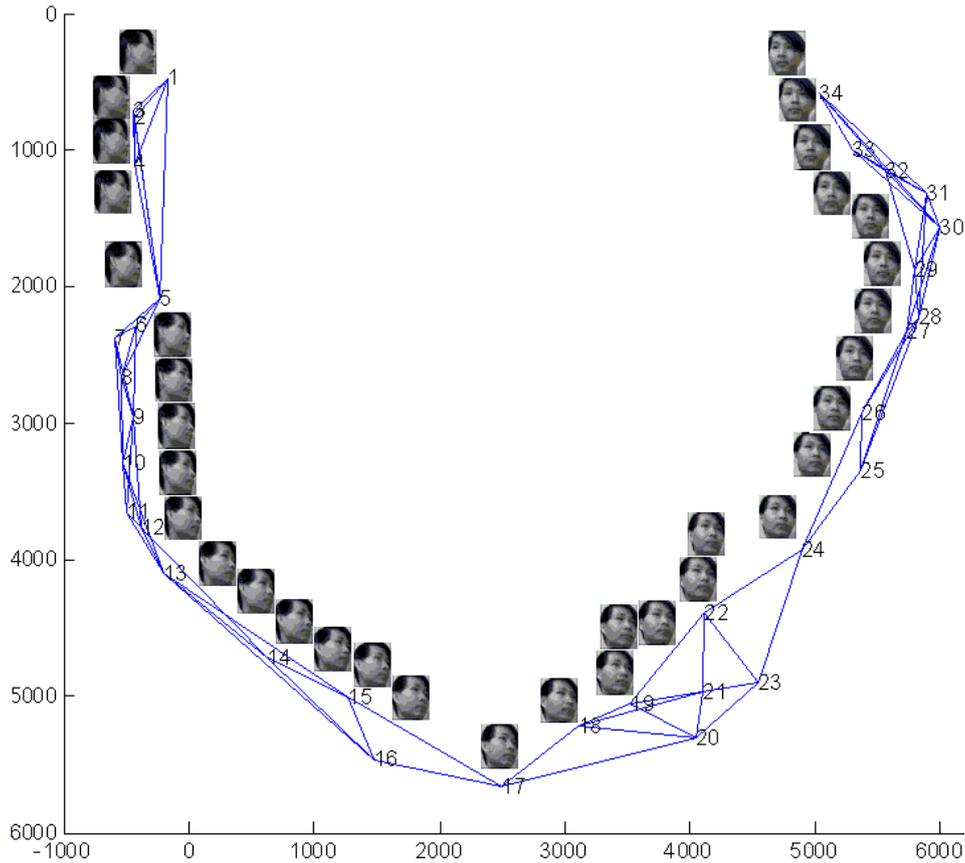

Fig. 23 The final dimension reduction result for the image sequence in Fig. 22

### 7. Conclusion and Discussion

In this paper, a novel dimension reduction method is proposed by the autonomous deforming (self-evolution) of data manifolds. The deforming is guided by the proposed deforming vector field including two kinds of interactions between data points. The elastic interaction preserves the topological structure of the data manifold, while the repelling interaction stretches and spreads the manifold. The flattening of the manifold in $R^n$ can be achieved as an emergent result of data point interactions. To overcome the problem caused by non-uniform sampling and "short-circuit edge", the soft neighborhood is proposed as an adaptive way to determine the interactions between neighbor points.

With analogy to the attractor in differential dynamic systems, the flattening effect of the proposed deforming vector field can be interpreted as the evolution of the data manifold to an "attracting state". If all the data points are in a same low-dimensional hyper-plane in $R^n$, the attracting or repelling vectors between any two of them are all in the same hyper-plane. Moreover, the displacement vectors of the data points under such attracting and repelling interactions are also within the same hyper-plane, and no point will move out of this low-dimensional hyper-plane. Thus it can be regard as "attracting state". The evolution of the data manifold under the proposed deforming field will approach such attracting state (i.e. the manifold will be flattened).

The experiment results on cylindrical surface and Gaussian surface prove that the proposed method can effectively flatten the two typical surfaces of the bending or concave-convex case. Other experiments have been carried out on real-world data sets, including the object images with changing size and angle of view, face images with changing illumination angle and intensity, and also face image sequences captured when the subjects turn their heads. The experimental results prove that effective dimension reduction can be achieved by the proposed method, the intrinsic dimension can be revealed, and there are meaningful interpretations for each dimension after DR. Moreover, it has the potential to reveal implicit feature in the data set. Further study will investigate detailed characteristics of the final stable shape of the deforming manifold, and its relationship between the algorithm parameters (i.e. weight coefficients), which may provide clues for new method design in dimension reduction.